\pdfoutput=1

\documentclass[11pt]{article}

\usepackage{rotating} 

\usepackage[final]{acl}

\usepackage{times}
\usepackage{latexsym}
\usepackage{adjustbox} 
\usepackage{booktabs} 
\usepackage{tabularx} 
\usepackage{array}    
\usepackage{caption}
\usepackage[table,xcdraw]{xcolor} 
\usepackage{float}

\usepackage{microtype}
\usepackage[T1]{fontenc}

\usepackage[utf8]{inputenc}

\usepackage{inconsolata}

\usepackage{graphicx}

\title{Paraphrasing Attack Resilience of Various AI-Generated Text Detection Methods 
 }

\author{
  Andrii Shportko \\
  Northwestern University \\
  Evanston, IL, USA \\
  \texttt{andre.s@u.northwestern.edu} 
  \And
  Inessa Verbitsky \\
  Northwestern University \\
  Evanston, IL, USA \\
  \texttt{inessa@u.northwestern.edu} 
}

\begin{document}
\maketitle
\begin{abstract}

The recent large-scale emergence of LLMs has left an open 
space for dealing with their consequences, such as plagiarism or the spread of false information on the Internet. Coupling this with the rise of AI detector bypassing tools, reliable machine-generated text detection is in increasingly high demand. We investigate the paraphrasing attack resilience of various machine-generated text detection methods, evaluating three approaches: fine-tuned RoBERTa, Binoculars, and text feature analysis, along with their ensembles using Random Forest classifiers. We discovered that Binoculars-inclusive ensembles yield the strongest results, but they also suffer the most significant losses during attacks. In this paper, we present the dichotomy of performance versus resilience in the world of AI text detection, which complicates the current perception of reliability among state-of-the-art techniques.

\end{abstract}

\section{Introduction}

\hspace{\parindent} The widespread use of LLMs can be precarious when left unchecked, with the consequences ranging from intellectual dishonesty to the spread of fake news on social media.  \citet{ELALI2023100706} found that AI chatbots can easily produce both realistic-looking  academic results and a polished manuscript that may well be accepted to a conference and published. Since scientific research, especially medical, is often falsified, the emergence of such a possibility opens up a dangerous playing field \citep{Phogat2023}. It was found that 
14\% of scientists were aware of colleagues who falsified results, whereas 72\% of scientists knew of colleagues who engaged in questionable research practices \citep{Fanelli2009}. More incidents of AI being used in the case of fake news spreading on the internet can be found in the Ethics Statement.

What is particularly concerning about this is that humans have been found to perform rather poorly on manual detection of AI-written text. In particular, human performance has shown to be only marginally better than random classification \citep{wu2024surveyllmgeneratedtextdetection}. In fact, in a study involving over 130 subjects, \citet{Kumar_Mindzak_2024} found that participants were only able to correctly identify AI-generated text with an accuracy rate of 24\%. Concerning the use of AI in academia, \citet{Gao2022.12.23.521610} conducted an experiment where participants were to identify whether abstracts for academic papers were written by ChatGPT or a human. They found that only 68\% of the AI-detected abstracts were correctly classified. Such a precedent makes a strong case for the necessity of precise automated AI text detection mechanisms. 

With the emergence of freely accessible sites such as ZeroGPT, DetectGPT, and Quillbot, bypassing attacks have been developed against these technologies. Methods which are commonly used include automated paraphrasing tools, prompt engineering, and the calculated addition of errors into AI-generated text \citep{Perkins2024}. It has been generally shown that the use of these methods decreases the efficacy of the detection tool; however, we aim to put together a more comprehensive analysis of the leading AI detection methods against bypassing attacks. In this paper, we will focus on paraphrasing attacks. 

The leading state-of-the-art detectors can be categorized into two paradigms, those using training-based and training-free mechanisms \citep{wang2025genai}. Most training-free approaches rely on statistical feature analysis and commonly look at perplexity, log probability, and n-grams \citep{chakraborty2023possibilitiesaigeneratedtextdetection}. Although training-based models have been widely leading, a recently developed methodology – Binoculars – proves successful in a zero-shot context, which stands out over multiple metrics \citep{hans2024spottingllmsbinocularszeroshot}. This approach is developed further in the Related Work section. Training-based methods largely rely on transformer models, namely RoBERTa \citep{DBLP:journals/corr/abs-1907-11692}, a masked-based model, easily fine-tunable for downstream tasks such as text classification. 

The methods we stacked to develop our own model include Binoculars, RoBERTa, and text feature analysis, which we justify due to their leading benchmarks (detailed in Related Work). 

\section{Related Work}

\subsection{Binoculars}
\hspace{\parindent}The Binoculars method relies on calculations from two closely related LLMs. It has a significant advantage over other SOTA methods in that it uses no training from the LLM that it is being tested on. This is significant, considering Binoculars still manages to surpass every open-source model that detects ChatGPT. Because other detectors rely on pretraining of the models they then test, the results fail to generalize when tested across multiple AI models. The Binoculars method, however, achieves high performance on a variety of datasets, which gather texts from different LLM sources. Furthermore, Binoculars addresses what they call the “Capybara Problem”, which in essence refers to the phenomenon of an LLM generating high-perplexity text simply due to a high-perplexity prompt being used. Other models which focus on raw perplexity will fail in such cases. Binoculars has an accuracy rate above 90\%, and a false positive rate of 0.01\%, using datasets which include\textit{ Writing prompts, News, and Student essays} \citep{verma-etal-2024-ghostbuster}.

\subsection{Text Features}
\hspace{\parindent} \citet{Munoz-Ortiz2024} analyzed linguistic patterns in human and LLM text to determine which features would provide for the most robust detection mechanism. Using extensive data from six different LLMs, including Llama and Falcon 7-b, they found that human writing tends to have less uniform sentence length distribution than AI. This conclusion is supported by \citet{Desaire2023}, who found that the standard deviation of sentence length was an important identifier in text classification. As one of our five text features used, we thus implement standard deviation of sentence length.

\subsection{Ensembling}
\hspace{\parindent}\citet{abburi2023generativeaitextclassification} analyze the success in using ensemble approaches for text classification. Their ensemble involves stacking DeBERTa, RoBERTa, and xLM-RoBERTa, fine-tuning each model for the appropriate tasks. They found that this approach reached 5th place in the English task and first place in the Multilingual for the Automated Text Identification shared task.

In fact, ensembling was highly used in Task 1 of the \textit{COLING 2025 GenAI Text detection workshop}, from which we use the dataset provided to train and evaluate our own model \citep{wang2025genai}. 
\citet{mobin2025luxverigenaidetectiontask}, whose approach scored 4th among contestants, relied on ensembling RoBERTa-base with other pre-trained transformer models. Our methodology also relies on RoBERTa, however, we ensemble it with Binoculars and text feature analysis, as justified above. 

\subsection{Bypassing}
\begin{figure*}[ht] 
    \centering
    \includegraphics[width=0.8\textwidth]{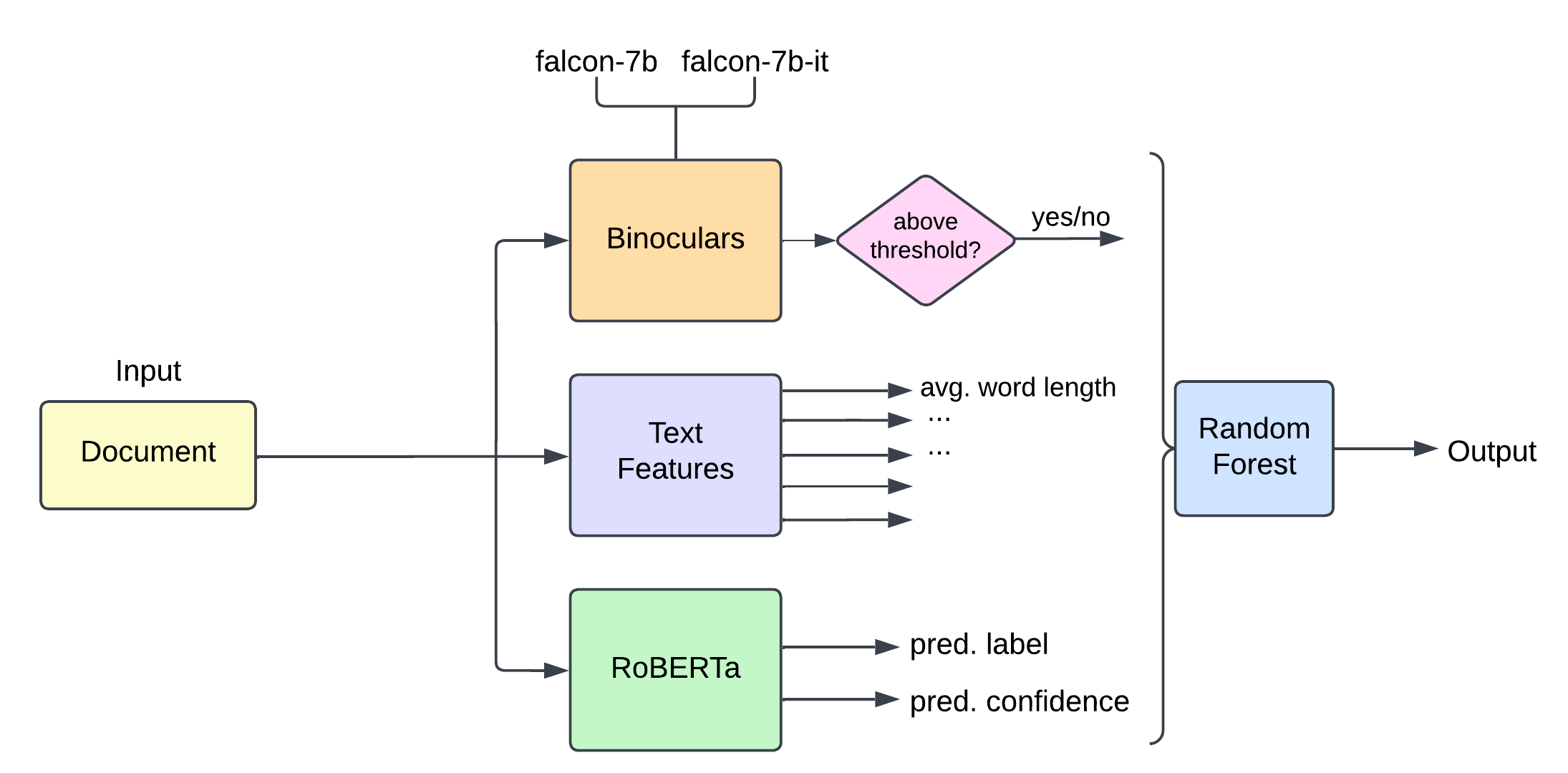} 
    \caption{Pipeline of our model}
    \label{fig:1}
\end{figure*}

\hspace{\parindent}The most prominent AI text-detection models relying on transformer fine-tuning have been tested against bypassing and proven to largely withstand it. \citet{krishna2023paraphrasingevadesdetectorsaigenerated} provide a critical baseline by demonstrating that controlled paraphrasing can significantly undermine the performance of AI-generated text detectors while maintaining semantic integrity. Their work, through the DIPPER model, shows that even minimal paraphrasing -- changing wording and sentence structure -- can drop detection accuracy significantly.

Some common AI detectors saw decreases of around 17\% in accuracy \citep{Perkins2024} when bypassing methods were employed. However, some more recently developed models were created specifically to withstand such attacks, such as the RADAR model \citep{hu2023radarrobustaitextdetection}, which trains the detector on paraphrasing schemes and achieves over 31.64\% of additional accuracy compared to previous methods. The Binoculars method, however, has not been tested against bypassing, thus its general efficacy remains unclear. This concern is explored in our paper. 

\section{Data and Methodology}

\setlength{\parskip}{0pt} 
\setlength{\parindent}{1em} 
\hspace{\parindent}To track the progress on machine-generated text detection, we use the materials of the competition on\textit{ Detecting AI Generated Content @COLING 2025 Task 1: Binary Machine-Generated Text Detection} \citep{wang2025genai}. It is an aggregation of other datasets that have been studied before, such as M4GT. The experiments in the following sections are based on the testing dataset that the final leaderboard used. All models use the training dataset, which is described in Appendix A.0.1.

First, we chose to fine-tune RoBERTa for AI text detection because it provided a substantial improvement in the model's ability to understand nuanced language differences. In essence, we added a final layer of size 2 for binary classification. It is also a well-tested approach in machine-generated text detection \citep{DBLP:journals/corr/abs-1907-11692}. We performed fine-tuning over a subset (12k entries) of the training set provided by the workshop. The hyperparameters are learning rate = $2e-5$, batch size = $16$, epochs = $4$, and training size = $20,000$, with a train/test split of $0.8$.

Second, we also test the Binoculars method, which reaches high accuracy and low false-positive rates over multiple LLM tested on, without relying on training data. Binoculars uses two closely related LLMs – '{\tt tiiuae/falcon-7b}', '{\tt tiiuae/falcon-7b-instruct}' – to calculate cross-perplexity, meaning perplexity is calculated using the log perplexity of text generated by one LLM and the next-token prediction of another. 

Third, we measured several document metrics that are related to AI detection. We selected 5 text markers: average word length, lexical diversity, punctuation frequency \citep{app13137901}, standard deviation of sentence length, and stopword ratio \citep{10.1145/3664476.3670465}. The selection of features was based on the entropy values from the Random Forests.

We combined the features extracted from each approach into a single vector for each text sample and fed it to the Random Forests model that acts like a meta-learner. This vector includes the prediction probabilities from the fine-tuned RoBERTa model as well as the predicted labels, the cross-perplexity scores from Binoculars, and the five document metrics we selected (Fig \ref{fig:1}). In the following sections, we will show the performance of all 7 different stackings of the models. 

Since we had limited resources, we manually chose $201$ random entries from the evaluation dataset, with the AI label, and fed them to the high-performance AI text detector bypasser GPTinf. We concatenated these paraphrased entries with $201$ randomly selected human-written entries from the same evaluation dataset. GPTinf claims to bypass all AI detectors, including Turnitin AI Detector, GPTZero, ZeroGPT, and GPTRadar. The dataset is published now on HuggingFace at '{\tt antebe1/paraphrased\_AI\_text}'.

Although the algorithm used by GPTInf is not publicized, their website states that it works by paraphrasing the inputted text--removing common phrasing and diversifying sentence structure by varying the wording, grammar, and ordering of words used \citep{GPTInf}. To calculate the confidence interval (CI) for the F1 score on the full dataset, we used a bootstrapping approach ($9000$ out of $73$k). To verify whether the differences between modules tested on were significant, we ran $21$ pair-wise McNemar statistical tests (Table \ref{tab:f1_comparison}). The Bonferroni correction for $\alpha=0.1$ is $0.0048$.

\section{Results}
\subsection{Binoculars}
\subsubsection{Observations}
\hspace{\parindent}For rapid testing purposes, all tests on Binoculars have been run on the devtest split of the dataset. 

\begin{figure}[t]
    \centering
    \includegraphics[width=\linewidth]{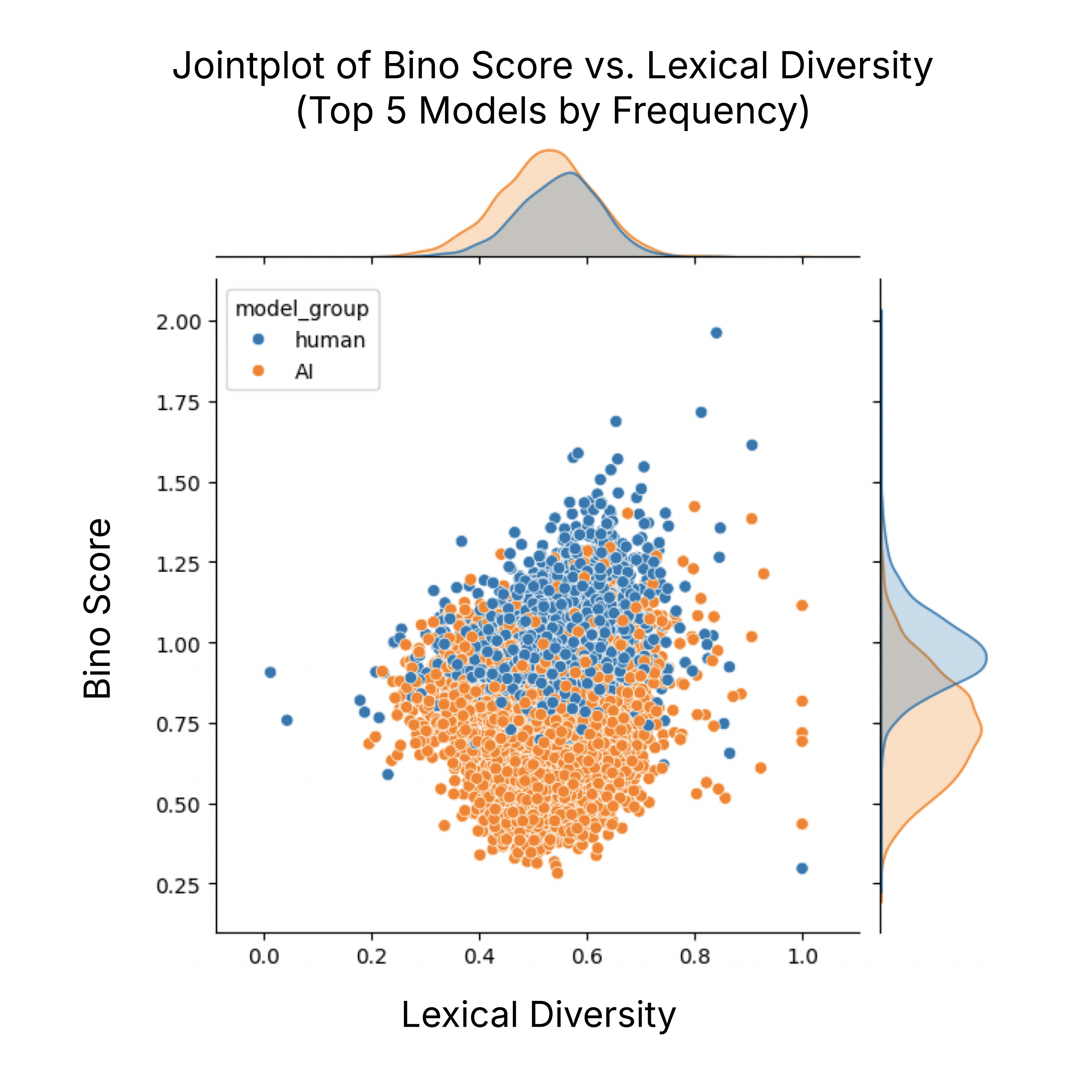}
    \caption{Binoculars results}
    \label{fig:Binoculars results}
\end{figure}

\begin{figure*}[t]
    \centering
    \includegraphics[width=0.7\linewidth]{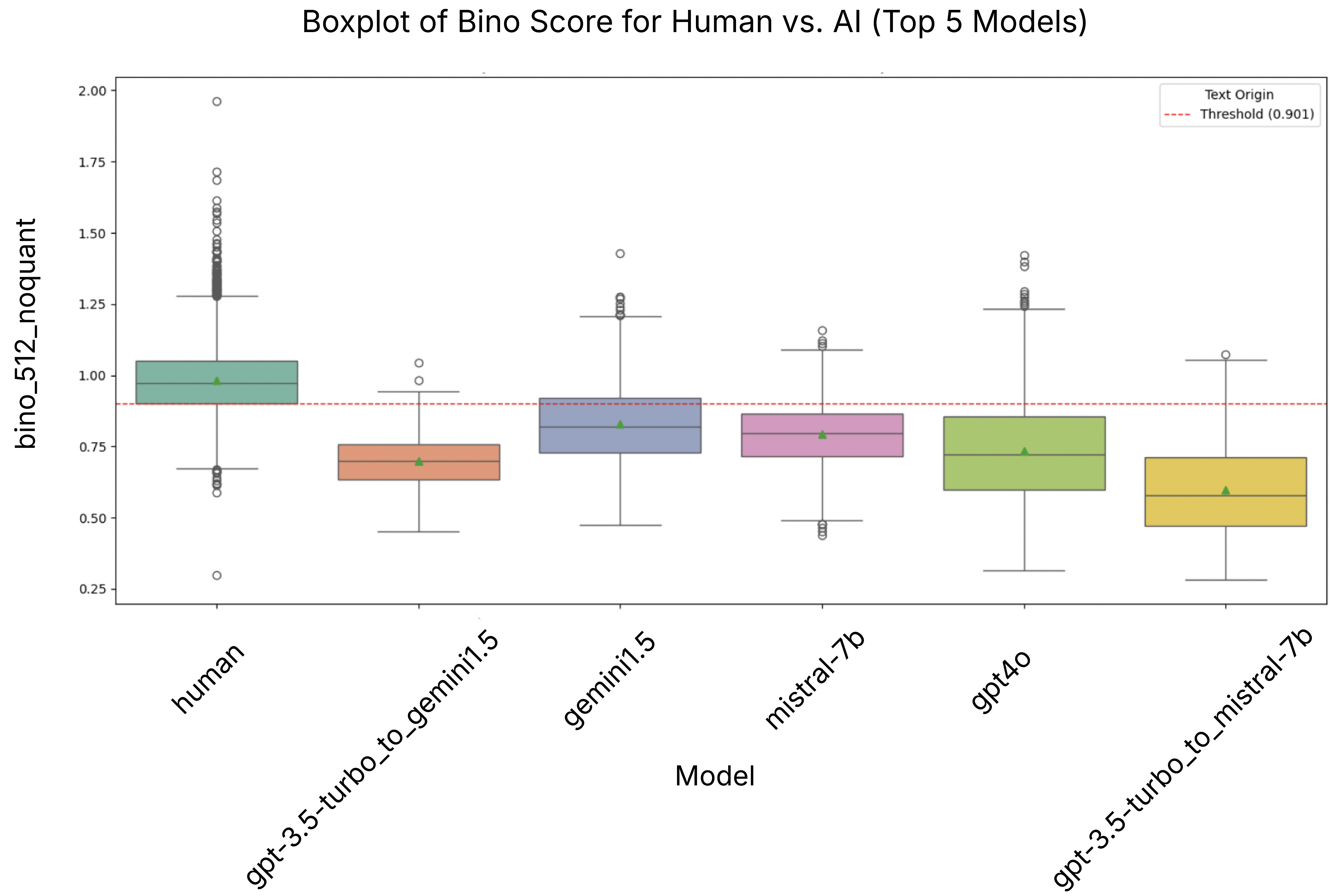}
    \caption{Binoculars score over context window of 512 w/o quantization}
    \label{fig:context-window}
\end{figure*}

\subsubsection{Context Window Effect}

\hspace{\parindent}We observed that the information gain increases as the context window increases. However, the information gain plateaus somewhere after $256-512$ tokens. The Jensen-Shannon (JS) divergence score (Fig \ref{fig:context-window}, see Appendix A.0.2), which measures the similarity between probability distributions, demonstrated significant improvements from $0.0373$ (context window size = $32$) to $0.2843$ (context window size = $512$). The JS score highlighted distinct effects between human-authored and AI-generated text as the context window increased.

\hspace{\parindent}The Binoculars score analysis reveals a clear separation between human and AI-generated text. 
Human-written content maintains the highest median score around $1.0$ (Fig \ref{fig:Binoculars results}) as predicted by the Binoculars paper, exhibiting notable variance and outliers. The critical threshold value of $0.901$, just as reported in the original paper, serves as a discriminator between human and AI-generated content. 

\subsection{Module Ensemble Comparisons}
\begin{figure}[H]
    \centering
    \includegraphics[width=\linewidth]{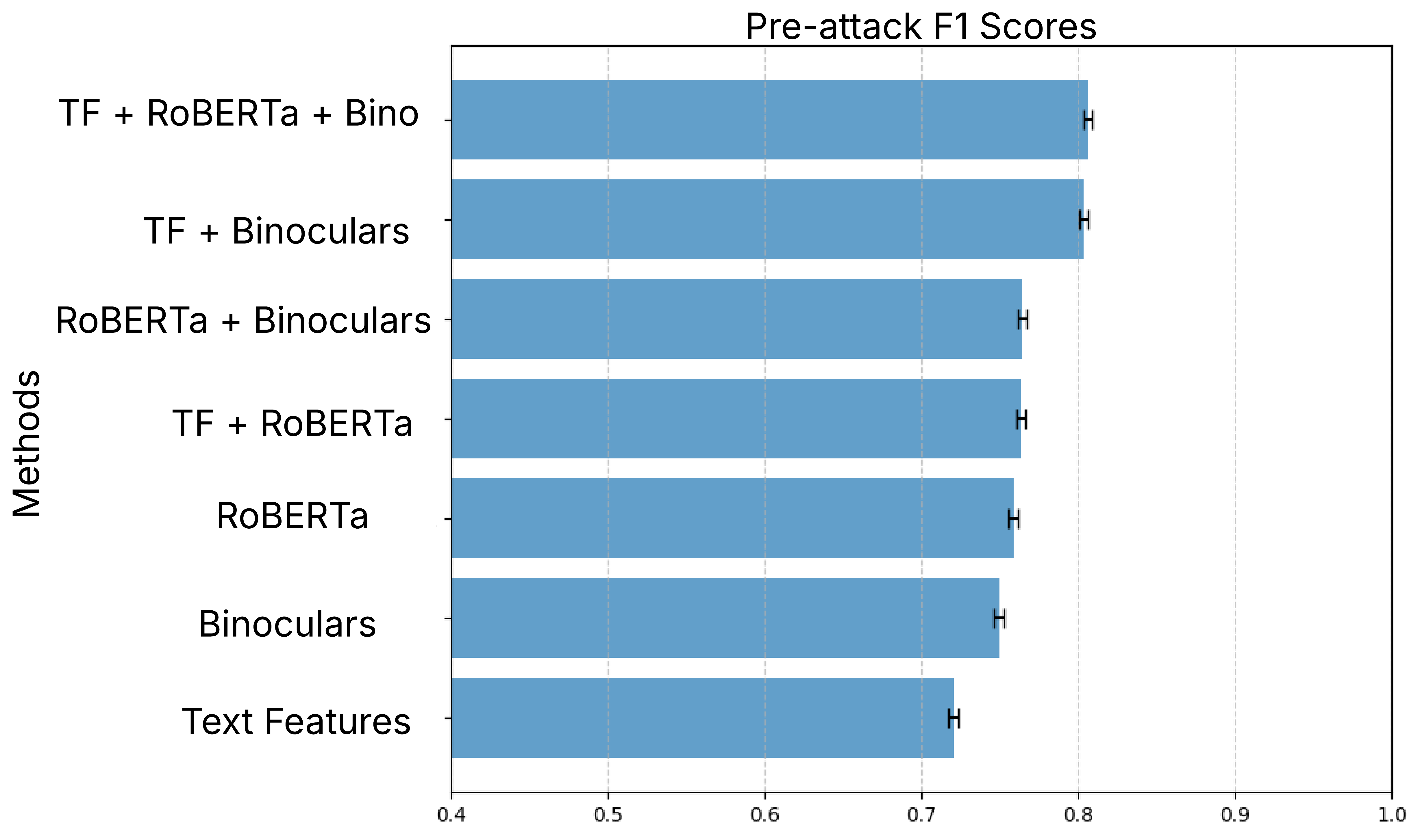}
    \caption{Pre-attack F1 scores}
    \label{fig:preattack accuracy}
\end{figure}

\hspace{\parindent}Three different modules give rise to 7 different ways to assemble them (Fig \ref{fig:preattack accuracy}). 

 The ensemble incorporating all modules (Text Features, RoBERTa, and Binoculars) achieves the highest F1 score of 80.61\%. The second-best performance is observed when Text Features and Binoculars are combined. While combining Text Features with RoBERTa or RoBERTa with Binoculars also improves performance compared to individual features, they fall short of the comprehensive ensemble. Notably, individual feature sets such as Text Features, RoBERTa, or Binoculars alone yield lower F1 scores than any combination of them (as seen in Table \ref{tab:sorted_models}). 
 

\subsection{Paraphrasing Attack}

\begin{figure}[H]
    \centering
    \includegraphics[width=\linewidth]{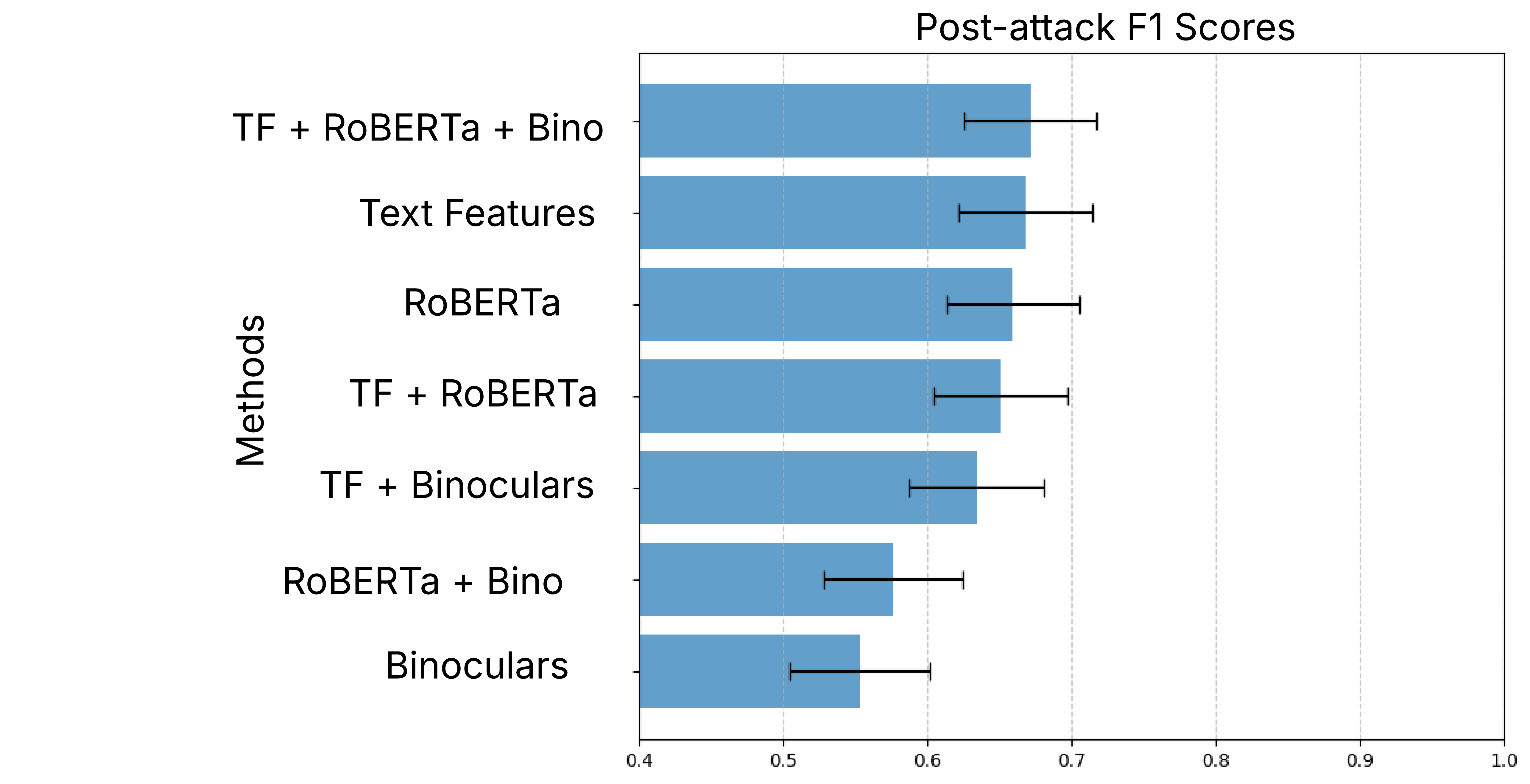}
    \caption{Post-attack F1 scores}
    \label{fig:enter-label}
\end{figure}

\hspace{\parindent}  Among individual models, RoBERTa demonstrated the highest resilience to paraphrasing attacks, showing almost no degradation (Table \ref{tab:sorted_models}). In contrast, the Binoculars method exhibited the most vulnerability, resulting in a significant degradation of $0.196$. 

Interestingly, the Text Features approach also showed almost no degradation in performance against paraphrased samples. The ensemble combining Text Features, RoBERTa, and Binoculars achieved the highest initial F1 score of $0.8061$ but experienced a notable drop in performance when faced with paraphrased samples, decreasing to $0.6716$. These findings highlight the varying degrees of resilience among different approaches to machine-generated text detection. RoBERTa's robustness suggests that its language understanding capabilities allow it to detect AI-generated text even after paraphrasing. The significant drop in Binoculars' performance indicates that its cross-perplexity approach may be more sensitive to changes in text structure and wording introduced by paraphrasing.

\begin{table}[ht]
\centering
\caption{F1 score drop per ensemble}
 \label{tab:sorted_models}
\renewcommand{\arraystretch}{1.2} 
\setlength{\tabcolsep}{4pt}       
\begin{adjustbox}{width=0.48\textwidth, center} 
\begin{tabular}{|>{\raggedright\arraybackslash}p{4.5cm}|c|c|c|}
\hline
\rowcolor[HTML]{EFEFEF} \textbf{Model}                          & \textbf{Initial F1} & \textbf{Paraphrased F1} & \textbf{Degradation} \\ \hline \hline
Binoculars                              & 0.7497                  & 0.5533                        & 0.1964                \\ \hline
RoBERTa + Binoculars                    & 0.7644                  & 0.5765                        & 0.1879                \\ \hline
Text Features + Binoculars              & 0.8035                  & 0.6340                        & 0.1695                \\ \hline
Text Features + RoBERTa + Binoculars    & 0.8061                  & 0.6716                        & 0.1345                \\ \hline
Text Features + RoBERTa                 & 0.7636                  & 0.6511                        & 0.1126                \\ \hline
Text Features                           & 0.7207                  & 0.6682                        & 0.0526                \\ \hline
RoBERTa                                 & 0.7586                  & 0.6594                        & 0.0992                \\ \hline
\end{tabular}
\end{adjustbox}
\end{table}

\begin{figure}[H]
    \centering
    \includegraphics[width=\linewidth]{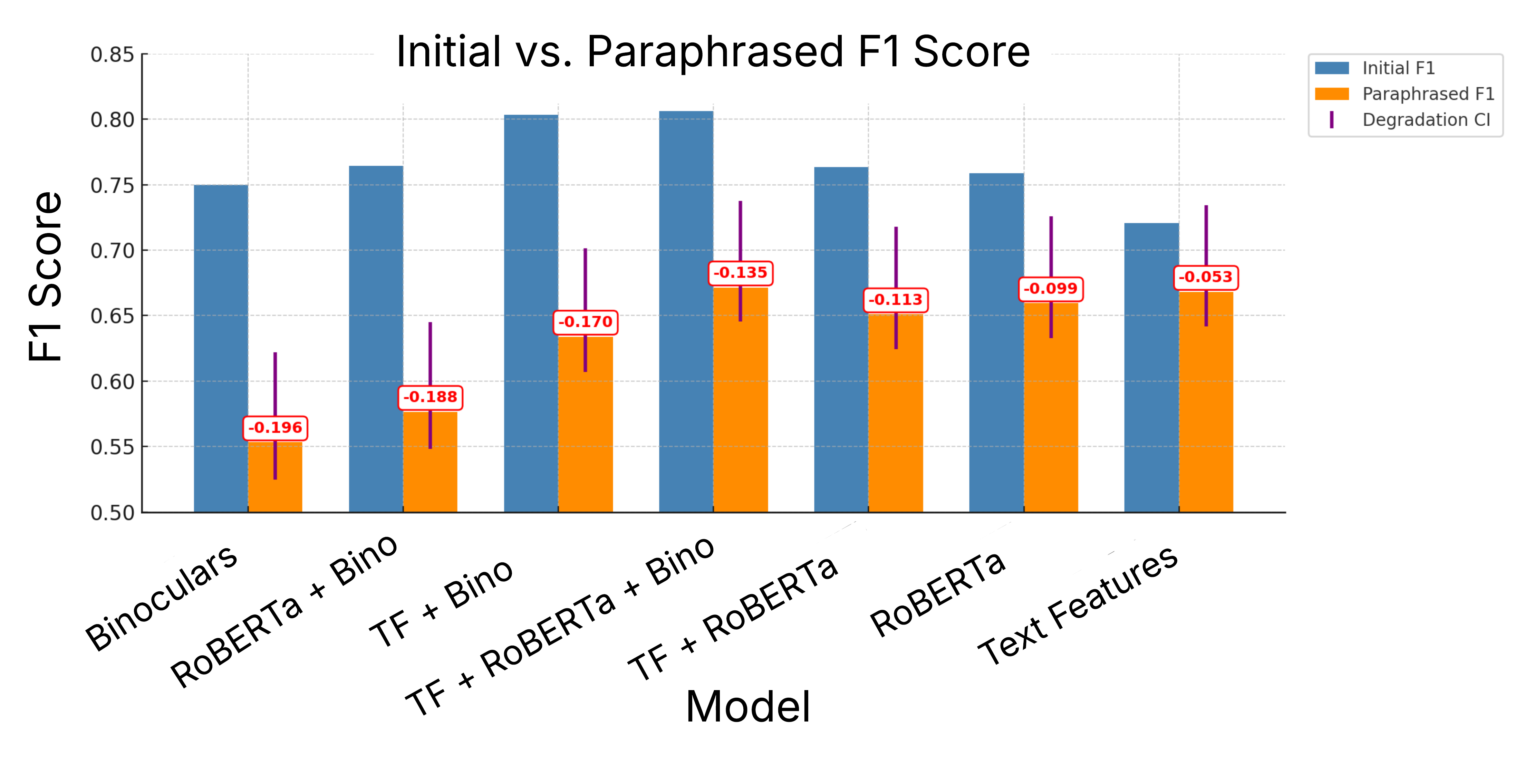}
    \caption{F1 score comparison and degradation}
    \label{fig:enter-label}
\end{figure}

\section{Discussion}
\subsection{Analysis of Results}
\hspace{\parindent}As demonstrated by our Results, we introduced a Cohesive Testing Framework (CTF) for classifying text as human- versus machine-written. Our system streamlines the ensembling process by feeding the document input into three detectors – Binoculars, Text Features, and RoBERTa, which are then stacked and evaluated by our meta-learner, Random Forest (as demonstrated in Fig \ref{fig:1}).  Our method employs 7 ways to combine 3 modules and make cross-comparisons, which allows for 1-to-1 comparisons between performance of modules. Our ensemble method proved significant information gain which outperforms many SOTA detectors. Namely, it would place us 4th on the COLING2025 Workshop leaderboard. 

Our second main result was our finding that the highest performing AI detectors had the worst results when it came to paraphrasing attacks. In fact, any ensemble that used Binoculars saw a significant decrease in F1 score. This is particularly interesting, as it reflects more generally “The Bitter Lesson” paradox – it seems that for every interpretable training-free method there is a better black-box approach.

\subsection{Future Work}
\hspace{\parindent}We suggest future works to build off our model by addressing the limitations we have laid out on the following page, as well as evaluating the detectors we looked at on different bypassing attacks, not only paraphrasing. Additionally, the methods evaluated were not tested for out-of-distribution prompts. Hense, accounting for this may add to a more comprehensive review of SOTA detectors. Sentiment analysis has been shown to be distinguishable between human and machine-written text, thus including this as a feature may contribute to some interesting results as well. 

\section{Conclusion}
\hspace{\parindent}We believe that the tradeoff between performance and resilience is significant enough to become a leading theme in the AI-detection community. For example, the reported high performance of  Binoculars on flagging Machine-generated text has suffered the most drastic loss under paraphrasing attacks. Under our testing framework, we also reaffirmed the significant information gain provided by the stacking of multiple detectors.  
\section*{Acknowledgments}
We thank the NU CS Microgrant for the computational units provided and Prof. Wood-Doughty for guidance and consultation on statistical analysis. The study resulting in this paper was assisted by a Conference Travel Grant from the Office of Undergraduate Research administered by Northwestern University's Office of the Provost. However, the conclusions, opinions, and other statements in this paper are the author's and not necessarily those of the sponsoring institution.
\clearpage

\section*{Limitations}

\hspace{\parindent}Our paraphrased dataset has $200$ entries, as we were unable to gain API access to the platform we used. Thus, although statistically significant, it should be important to replicate our results with a larger dataset. Additionally, we only tested paraphrasing generated by GPTInf, which may not capture the maximum extent of paraphrasing attack capabilities.

\section*{Ethics Statement}
\hspace{\parindent}When ChatGPT was released in 2022, it was widely unheard of and thus not largely anticipated, but within a short time frame, its popularity surged. The world had not been expecting such a capable and easily accessible system, and thus its use in academic settings by students, across the internet by scammers, and in almost every practical field by workers, skyrocketed within a very brief amount of time. As a result, the consequences of such wide-spread AI use have not been thoroughly accounted for, and recent studies of its very real and threatening possible repercussions have only begun to be released. It is then instrumental to first, study the effects of large-spread AI use, and following this, develop methods that can detect the use of AI, namely in writing.

The use of deepfakes have become increasingly prevalent in recent years. \citet{TRANDABAT20233822} investigated and assessed the threat of AI being used to generate deepfakes on a mass scale to be then published across the internet. Google published the DeepDream algorithm in 2015, which used a convolutional neural network, trained on millions of images, to first identify objects within images, and then using these patterns create an image corresponding to a requested object (for instance, an animal) from memory \citep{Miller2020}. Although the images that this network could produce were far from accurate and often combined elements of different objects from its training data, the release of DeepDream instantly sparked a race to use this technology and create something more powerful, as this was the first time deep learning was used to generate images from scratch. Soon, more models and algorithms were developed, which were more advanced, with time, shrinking the gap between human recognizability of what is evidently machine-generated in comparison to human-created. In their paper, \citet{TRANDABAT20233822} use this background to focus on the present-day role of AI across the internet, notably what is commonly referred to as “fake news”. They test a few classifiers on both human and AI-generated fake news, including RoBERTa. They find that the true positive rate of AI-detector models, such as RoBERTa, on AI-generated fake news is only 3\% higher than when run on human-generated fake news, thus making AI-generated fake news very difficult to recognize and highly useful for publishing false information online.

In 2024, a German magazine \textit{Die Aktuelle} published an interview with a famous Formula One driver, Michael Schumacher, which was created entirely by the AI chatbot, Character.ai, upon which the magazine was sued by Schumacher’s family \citep{ESPN2024}. Schumacher has been out of the image of the public eye for almost a decade due to a brain-injury following a sports accident. His family has taken immense action to keep his life post-accident in private, thus the release of this article resulted in great turmoil on the family and misled readers all around the world.

Overall, the consequences of fake news becoming prevalent can be unimaginably dangerous. In South Korea, AI has been widely used to generate ads containing falsified information and promote the listing of fraudulent drugs and hormonal therapies for sale to the public on the internet \citep{Park2024}. Because the sale of these treatments over-the-counter have not been government-approved, many of the drugs listed have not been properly studied, meaning the health consequences that may arise from them are unclear, which is critically unsafe.
We thus strongly emphasize the need for reliability in AI text detection, highlighting the absolute necessity for AI text detectors that are able to \textit{bypass} bypassers, in order to combat these problems addressed above and promote transparency across the Internet, in all fields and aspects. 

\clearpage

\bibliography{custom}

\appendix

\section{Appendix}

\subsubsection{Dataset}
The training dataset contained a total of ~610k entries from HC3, M4GT, and MAGE. The test dataset contained a total of ~74k entries from CUDRT, IELTS, NLPeer, PeerSum, and MixSet.  We replicated 3 methods as well as their different ensembles over the Random Forest classifier and evaluated their performance on the MGTD testing dataset. 

\subsubsection{Quantization Effect}

\setlength{\parskip}{0pt} 
\setlength{\parindent}{1em} 

\begin{figure*}[t]
    \centering
    \includegraphics[width=0.7\linewidth]{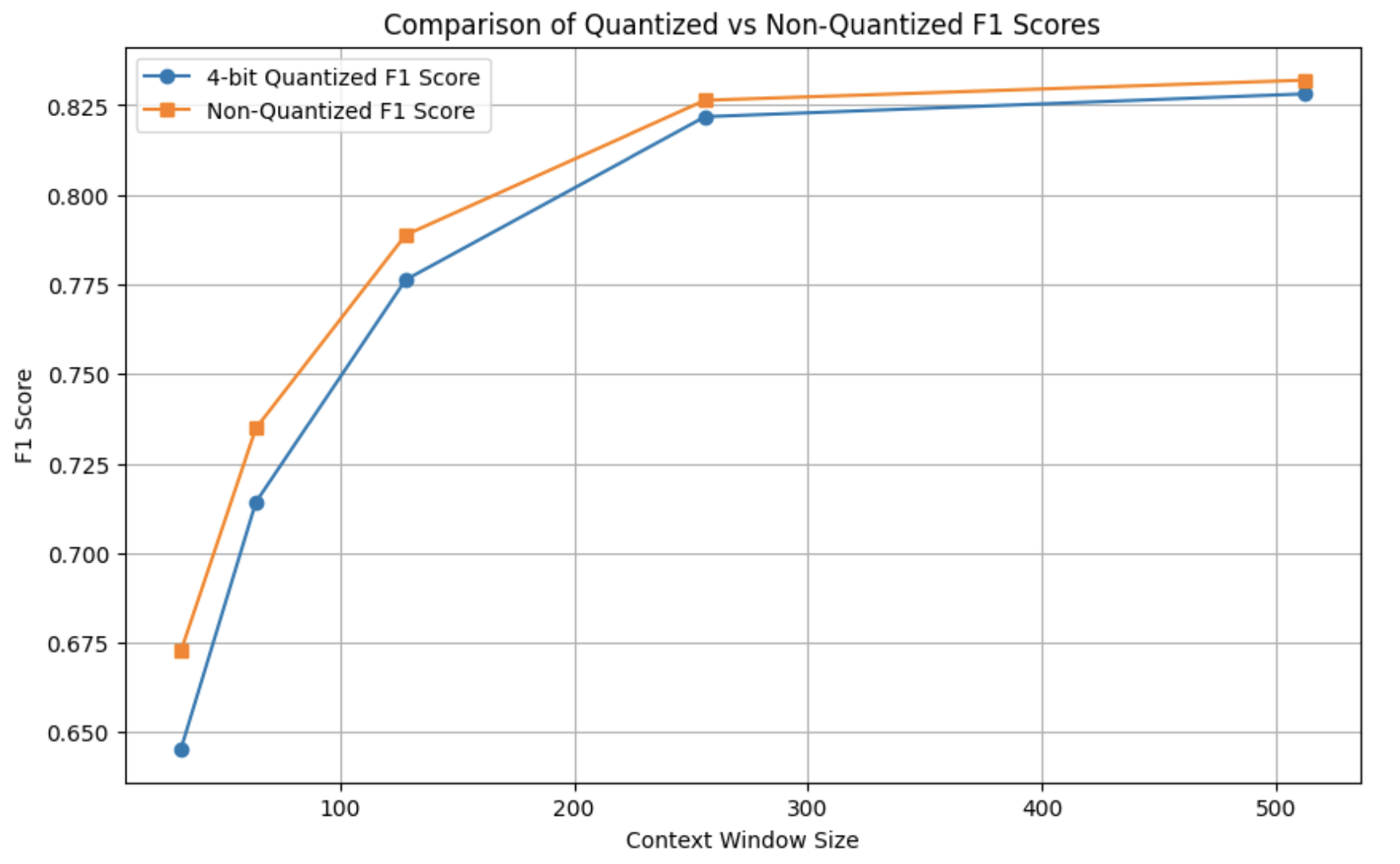}
    \caption{}
    \label{fig:quantization_effect}
\end{figure*}

\begin{figure*}[t]
    \centering
    \includegraphics[width=0.7\linewidth]{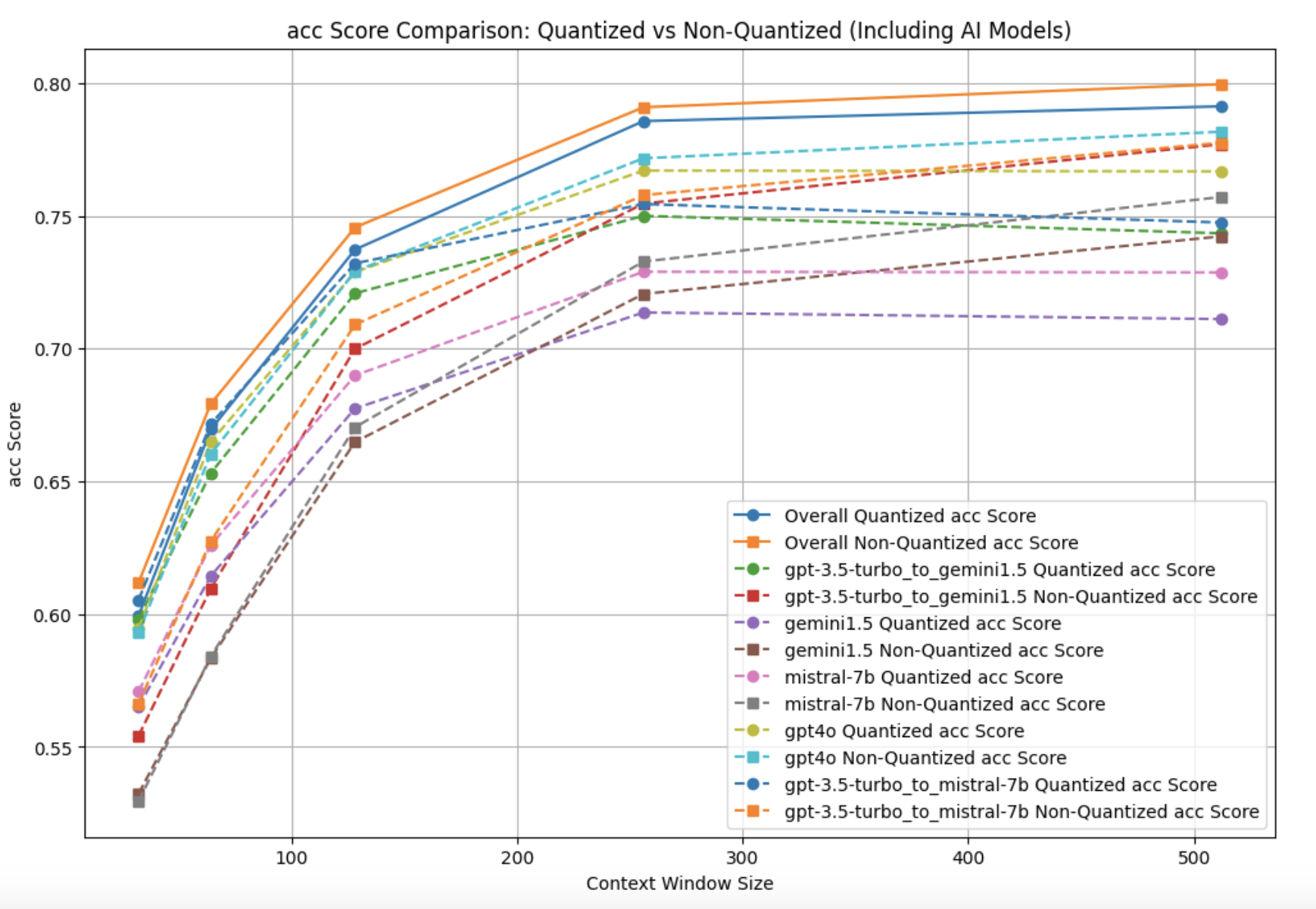}
    \caption{}
    \label{fig:quantization_multiple_models}
\end{figure*}
\hspace{\parindent}Quantization in machine learning is the process of reducing the precision of numerical values, typically converting floating point numbers to lower-bit representations, to decrease the model size and improve computational efficiency. In our experiments, we quantized HuggingFace “{\tt tiiuae/falcon-7b}” to replicate the 
paper.  The typical degradation effect was about ~2\% (Fig \ref{fig:quantization_effect}) and it was diminishing as context was increasing. This is unexpected because usually degradation effects for other tasks would be stronger. It took around ~27 GB of RAM to run “{\tt tiiuae/falcon-7b}” and “{\tt tiiuae/falcon-7b-instruct}” and 11 GB for 4-bit quantization of those models. We conclude that the marginal improvement of the F1 score is unimportant compared to the doubled Carbon Footprint. While non-quantized versions achieve better results, the marginal accuracy improvement must be weighed against the significantly higher computational requirements, particularly in resource-constrained environments.

\begin{figure*}[t]
    \centering
    \includegraphics[width=0.7\linewidth]{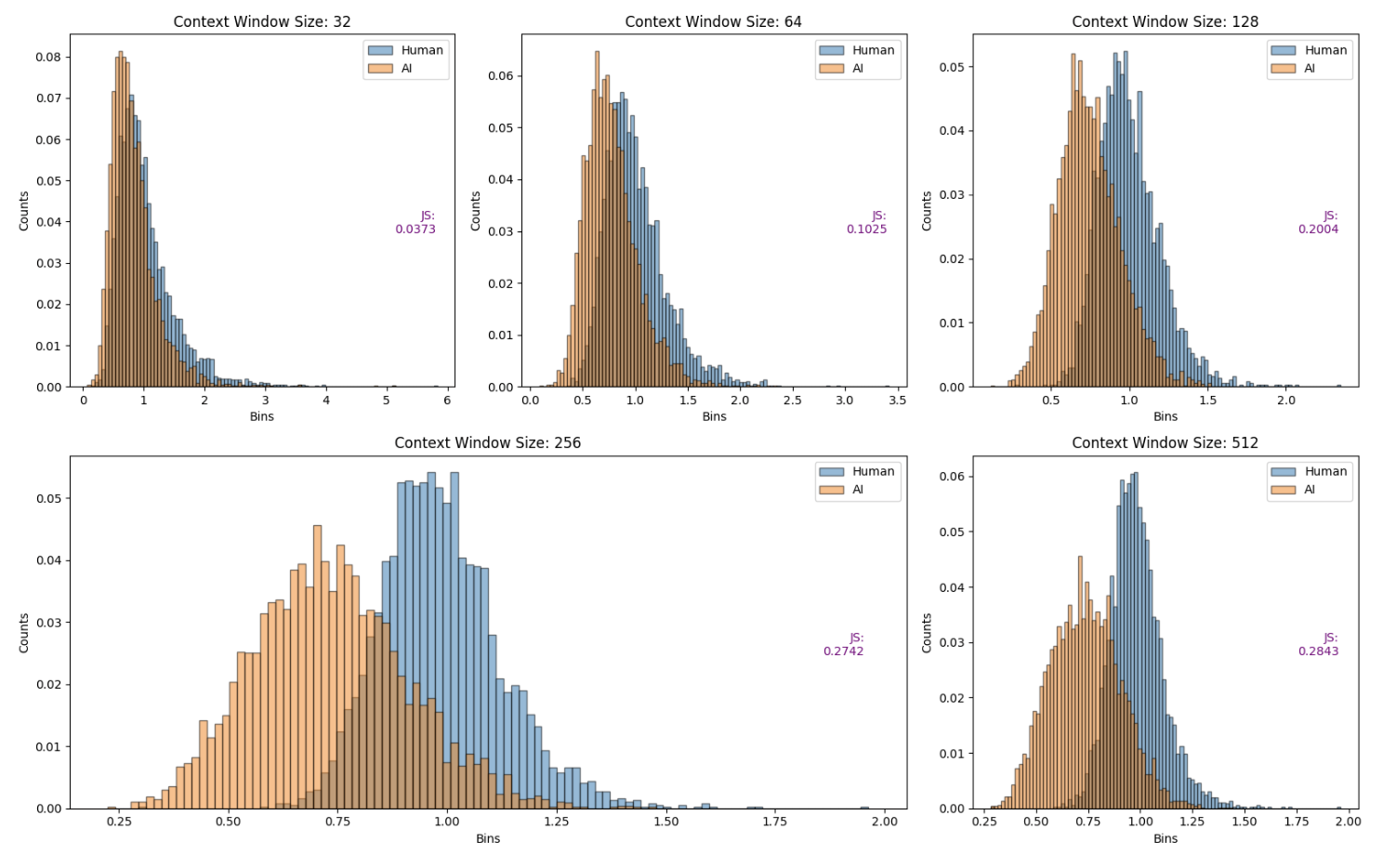}
    \caption{}
    \label{fig:context_window}
\end{figure*}

In Fig \ref{fig:quantization_multiple_models}, while some models are more resilient to Binocular detection (gemini1.5) than others (gpt4o), the trend is repeated for all models. Context window size significantly impacts detection accuracy, with substantial improvements observed as the window expands from 128 to 256 tokens. The optimal range lies between 256-512 tokens, though performance gains diminish notably beyond 300 tokens. The maximum accuracy peaks at approximately 0.80 for top-performing models at 512 tokens. Non-quantized models consistently demonstrate superior accuracy compared to their quantized counterparts, with approximately 2\% better performance.

\clearpage
\begin{table}[!ht]
    \centering
    \renewcommand{\arraystretch}{1.2}
    \setlength{\tabcolsep}{6pt}
    \begin{tabular}{|l|l|r|r|r|r|l|}
    \hline
        \rowcolor[gray]{0.9} \textbf{Module 1} & \textbf{Module 2} & \textbf{p-value} & \textbf{F1 (M1)} & \textbf{F1 M1} & \textbf{F1 Diff} & \textbf{Higher F1} \\ \hline
        Bino & TF + Bino + RoBERTa & 0.0166 & 0.5533 & 0.6716 & -0.1183 & TF + Bino + RoBERTa \\ \hline
        TF & Bino & 0.1310 & 0.6682 & 0.5533 & 0.1149 & TF \\ \hline
        Bino & RoBERTa & 0.2744 & 0.5533 & 0.6594 & -0.1061 & RoBERTa \\ \hline
        Bino + RoBERTa & TF + Bino + RoBERTa & 0.0478 & 0.5765 & 0.6716 & -0.0950 & TF + Bino + RoBERTa \\ \hline
        TF & Bino + RoBERTa & 0.2783 & 0.6682 & 0.5765 & 0.0917 & TF \\ \hline
        Bino & TF + RoBERTa & 0.2374 & 0.5533 & 0.6368 & -0.0835 & TF + RoBERTa \\ \hline
        RoBERTa & Bino + RoBERTa & 0.4397 & 0.6594 & 0.5765 & 0.0829 & RoBERTa \\ \hline
        Bino & TF + Bino & 0.0688 & 0.5533 & 0.6202 & -0.0669 & TF + Bino \\ \hline
        TF + RoBERTa & Bino + RoBERTa & 0.3924 & 0.6368 & 0.5765 & 0.0603 & TF + RoBERTa \\ \hline
        TF + Bino & TF + Bino + RoBERTa & 0.2809 & 0.6202 & 0.6716 & -0.0514 & TF + Bino + RoBERTa \\ \hline
        TF & TF + Bino & 0.5493 & 0.6682 & 0.6202 & 0.0480 & TF \\ \hline
        TF + Bino & Bino + RoBERTa & 0.2387 & 0.6202 & 0.5765 & 0.0436 & TF + Bino \\ \hline
        RoBERTa & TF + Bino & 0.4515 & 0.6594 & 0.6202 & 0.0393 & RoBERTa \\ \hline
        TF + RoBERTa & TF + Bino + RoBERTa & 0.0623 & 0.6368 & 0.6716 & -0.0348 & TF + Bino + RoBERTa \\ \hline
        TF & TF + RoBERTa & 0.4505 & 0.6682 & 0.6368 & 0.0314 & TF \\ \hline
        Bino & Bino + RoBERTa & 0.3751 & 0.5533 & 0.5765 & -0.0232 & Bino + RoBERTa \\ \hline
        RoBERTa & TF + RoBERTa & 0.5389 & 0.6594 & 0.6368 & 0.0226 & RoBERTa \\ \hline
        TF + Bino & TF + RoBERTa & 0.4853 & 0.6202 & 0.6368 & -0.0166 & TF + RoBERTa \\ \hline
        RoBERTa & TF + Bino + RoBERTa & 0.1097 & 0.6594 & 0.6716 & -0.0121 & TF + Bino + RoBERTa \\ \hline
        TF & RoBERTa & 0.4335 & 0.6682 & 0.6594 & 0.0087 & TF \\ \hline
        TF & TF + Bino + RoBERTa & 0.3719 & 0.6682 & 0.6716 & -0.0034 & TF + Bino + RoBERTa \\ \hline
    \end{tabular}
    \caption{Statistical Comparison of Pre-Attack F1 Scores Across Different Module Combinations (n\_bootstrap = $5000$, n\_subset = $73000$) }
    \label{tab:f1_comparison}
\end{table}
\clearpage
\begin{table}[!ht]
    \centering
    \renewcommand{\arraystretch}{1.2}
    \setlength{\tabcolsep}{6pt}
    \begin{tabular}{|l|l|r|r|r|r|l|}
    \hline
        \rowcolor[gray]{0.9} \textbf{Module 1} & \textbf{Module 2} & \textbf{p-value} & \textbf{F1 (M1)} & \textbf{F1 M1} & \textbf{F1 Diff} & \textbf{Higher F1} \\ \hline
        Bino & TF + Bino + RoBERTa & 0.0126 & 0.5533 & 0.6716 & -0.1183 & TF + Bino + RoBERTa \\ \hline
        TF & Bino & 0.1363 & 0.6682 & 0.5533 & 0.1149 & TF \\ \hline
        Bino & RoBERTa & 0.2675 & 0.5533 & 0.6594 & -0.1061 & RoBERTa \\ \hline
        Bino + RoBERTa & TF + Bino + RoBERTa & 0.0515 & 0.5765 & 0.6716 & -0.0950 & TF + Bino + RoBERTa \\ \hline
        TF & Bino + RoBERTa & 0.2706 & 0.6682 & 0.5765 & 0.0917 & TF \\ \hline
        Bino & TF + RoBERTa & 0.2255 & 0.5533 & 0.6368 & -0.0835 & TF + RoBERTa \\ \hline
        RoBERTa & Bino + RoBERTa & 0.4338 & 0.6594 & 0.5765 & 0.0829 & RoBERTa \\ \hline
        Bino & TF + Bino & 0.0682 & 0.5533 & 0.6202 & -0.0669 & TF + Bino \\ \hline
        TF + RoBERTa & Bino + RoBERTa & 0.4097 & 0.6368 & 0.5765 & 0.0603 & TF + RoBERTa \\ \hline
        TF + Bino & TF + Bino + RoBERTa & 0.2845 & 0.6202 & 0.6716 & -0.0514 & TF + Bino + RoBERTa \\ \hline
        TF & TF + Bino & 0.5374 & 0.6682 & 0.6202 & 0.0480 & TF \\ \hline
        TF + Bino & Bino + RoBERTa & 0.2311 & 0.6202 & 0.5765 & 0.0436 & TF + Bino \\ \hline
        RoBERTa & TF + Bino & 0.4425 & 0.6594 & 0.6202 & 0.0393 & RoBERTa \\ \hline
        TF + RoBERTa & TF + Bino + RoBERTa & 0.0635 & 0.6368 & 0.6716 & -0.0348 & TF + Bino + RoBERTa \\ \hline
        TF & TF + RoBERTa & 0.4581 & 0.6682 & 0.6368 & 0.0314 & TF \\ \hline
        Bino & Bino + RoBERTa & 0.3738 & 0.5533 & 0.5765 & -0.0232 & Bino + RoBERTa \\ \hline
        RoBERTa & TF + RoBERTa & 0.5420 & 0.6594 & 0.6368 & 0.0226 & RoBERTa \\ \hline
        TF + Bino & TF + RoBERTa & 0.4906 & 0.6202 & 0.6368 & -0.0166 & TF + RoBERTa \\ \hline
        RoBERTa & TF + Bino + RoBERTa & 0.1057 & 0.6594 & 0.6716 & -0.0121 & TF + Bino + RoBERTa \\ \hline
        TF & RoBERTa & 0.4461 & 0.6682 & 0.6594 & 0.0087 & TF \\ \hline
        TF & TF + Bino + RoBERTa & 0.3826 & 0.6682 & 0.6716 & -0.0034 & TF + Bino + RoBERTa \\ \hline
    \end{tabular}
    \caption{Statistical Comparison of Post-Attack F1 Scores Across Different Module Combinations (n\_bootstrap = $5000$, n\_subset = $402$)}
    \label{tab:f1_comparison}
\end{table}

\end{document}